# Assessing the potential of LLM-assisted annotation for corpus-based pragmatics and discourse analysis: The case of apologies

Danni Yu, Luyang Li, Hang Su, and Matteo Fuoli
School of European Languages and Cultures, Beijing Foreign Studies University | School of Information Science and Technology, Beijing Foreign Studies University | Center for Foreign Languages and Literature, Sichuan International Studies University | Department of English Language and Linguistics, University of Birmingham

## Abstract

Certain forms of linguistic annotation, like part of speech and semantic tagging, can be automated with high accuracy. However, manual annotation is still necessary for complex pragmatic and discursive features that lack a direct mapping to lexical forms. This manual process is time-consuming and error-prone, limiting the scalability of function-to-form approaches in corpus linguistics. To address this, our study explores the possibility of using large language models (LLMs) to automate pragma-discursive corpus annotation. We compare GPT-3.5 (the model behind the free-to-use version of ChatGPT), GPT-4 (the model underpinning the precise mode of Bing chatbot), and a human coder in annotating apology components in English based on the local grammar framework. We find that GPT-4 outperformed GPT-3.5, with accuracy approaching that of a human coder. These results suggest that LLMs can be successfully deployed to aid pragma-discursive corpus annotation, making the process more efficient, scalable and accessible.

## 1. Introduction

Annotation is a key aspect of contemporary corpus linguistics. Annotated corpora allow researchers to perform complex corpus queries, test hypotheses about language structure, and gain a better understanding of how language is used to communicate and interact across situational contexts (Leech, 1997). While certain linguistic attributes like part of speech can be annotated automatically with high accuracy (e.g. Garside and Smith, 1997), the analysis of pragmatic and discursive elements in corpora continues to rely heavily on manual annotation (e.g. Taylor, 2016, Cavasso & Taboada, 2021, Põldvere et al., 2022). This is mainly because these features frequently extend beyond the scope of individual lexical units and lack straightforward and definitive mapping onto specific lexical forms (Rühlemann & Aijmer, 2015). However, manual corpus annotation is a complex process that requires specialized skills, extensive training, and substantial time investment. It is also prone to human errors and inconsistencies, which can undermine accuracy and reliability. These challenges have hindered the scalability and widespread adoption of function-to-form corpus analysis of pragmatic and discursive phenomena. But what if we could use Artificial Intelligence to automate the corpus annotation process, drastically reducing the time and resources needed?

Recent advancements in AI driven by large language models (LLMs) –advanced machine learning models that use deep neural networks to process and learn from vast amounts of text data – have enabled significant improvements in automating complex language tasks such as text generation, translation, and question answering, achieving unprecedented levels of sophistication and accuracy. Researchers in the field of Natural Language Processing (NLP) have begun exploring the potential of using LLMs to assist the task of annotating corpora, achieving promising results across several applications (e.g., Ding et al., 2023; Frei & Kramer, 2023; Gilardi et al., 2023). However, it is unclear whether similar levels of performance can be achieved when annotating pragma-discursive features. This is because the annotation schemes used in NLP are significantly different from those commonly used in corpus linguistics. In NLP, these schemes are either designed for practical computational applications (e.g. automatically retrieving mentions of people, places, organizations etc.) or are based on analytical frameworks that are not widely adopted in corpus linguistic research (e.g., sentiment analysis). Currently, no studies within corpus linguistics have assessed the viability of employing LLMs to automate the process of annotating corpora. As a result, the potential of such technological advancements in aiding corpus-based pragmatics and discourse analysis remains unexplored.

This study, to our knowledge, is the first to investigate the possibility of using LLMs to automate pragma-discursive corpus coding. LLMs offer a significant opportunity to make this process substantially more efficient and scalable by reducing the amount of manual work required. In addition, a key benefit of these models is their user-friendly nature, eliminating the need for advanced programming skills. If successful, this approach has the potential to revolutionize the fields of corpus-based pragmatics and discourse studies by opening up new possibilities for large scale, function-to-form research. To assess the potential of LLM-assisted pragma-discursive corpus annotation, we compare the performance of two of the most advanced LLMs developed to date – GPT-3.5 (implemented in the free-to-use version of ChatGPT chatbot) and GPT-4 (which powers the precise mode of Bing chatbot) – and a human coder in annotating the functional components of apologies based on a *local grammar* approach (Su and Wei, 2018). We choose the speech act of apology to test our approach because it has been studied extensively in corpus linguistics and beyond and effectively showcases the typical challenges encountered in annotating the functional elements associated with such pragmatic functions. A key step in integrating LLMs into a language task is *prompt engineering*, which refers to the careful crafting of instructions or queries given to a language model, so as to elicit desired responses, in this case accurately annotated linguistic data. One of the purposes of the present study is therefore to develop a prompt design strategy that yields the highest possible level of accuracy in coding instances of apologies. By doing so, we offer a replicable protocol that could be either applied to the same annotation task or adapted to similar corpus annotation tasks.

## 2. Corpus annotation: long-standing challenges, new opportunities

Briefly, corpus annotation refers to the practice of adding detailed linguistic information to texts within a corpus to enable more focused and sophisticated analysis.

This section begins by discussing the challenges involved in this process, with a particular focus on the complexities of pragmatic and discourse-level annotation. Next, the discussion turns to consider the opportunities created by recent advancements in AI for this field.

**2.1** Challenges in automating pragmatic and discourse-level annotation

Corpora can be annotated at various levels, including phonetic, prosodic, grammatical, semantic and pragmatic/discursive [1] (Leech, 1993). One of the earliest and most common forms of corpus annotation is part-of-speech tagging, which involves labelling each word in a corpus with its corresponding grammatical category. Another common technique is parsing, which uses part-of-speech information to show how words relate syntactically (McEnery & Wilson, 2001: 53). Semantic tagging, which involves categorizing words into broad meaning categories, has gained traction in recent years thanks to the advancement of automated annotation software (e.g., Rayson et al., 2004). The metalinguistic 'tags' inserted in the corpus texts can serve a multitude of purposes, including refining corpus queries, investigating lexico-grammatical patterns of language use, validating and enhancing linguistic theories, aiding in the compilation of dictionary entries, and identifying predominant themes and discourses within corpora (Garside et al., 1997). Linguistic annotation thus substantially amplifies the capabilities of corpus tools and has for this reason evolved into an integral and indispensable component of corpus-based research.

Given the many benefits corpus annotation brings, substantial efforts have gone into creating annotation schemes for describing different linguistic aspects, as well as developing computational systems to automate this process. In an ideal scenario, software would be able to automatically tag features at all levels of linguistic description. However, that level of automation has not yet been achieved. Although certain forms of linguistic annotation, such as part-of-speech tagging, dependency parsing and semantic tagging can be performed automatically with high degrees of accuracy (McEnery & Hardie, 2012: 31), achieving the same level of success for other types of coding remains an elusive goal. Pragmatic and discourse-level features, in particular, present considerable challenge for automated analysis. This can be attributed to three main factors. First, pragma-discursive features often transcend the boundaries of individual lexical units, significantly complicating the annotation task as there are no consistent criteria for determining the units to be coded. For instance, the speech act of apology, which we use as a test case in our study, is rarely (if ever) only made up of the *illocutionary force indicating device* or IFID in English (e.g., *sorry* or *apologies*) (Blum-Kulka et al., 1989). Instead, apologies routinely incorporate explanations and offers of repair, the linguistic scope and complexity of which cannot be predicted in advance (e.g., Page, 2014; Su and Wei, 2018). Second, pragma-discursive functions are in most cases realized via an open-ended set of linguistic forms and lack a direct and unequivocal mapping onto specific lexical items. For instance, Lutzky & Kehoe (2017a) show that in online settings, apologies can be performed using less prototypical expressions like the word *oops*, which are not commonly considered in form-based research on apology. Similarly, feelings, attitudes and stances can be expressed in discourse through a wide variety of lexical forms across word classes like verbs (e.g., *love*), adjective phrases (e.g., *extremely talented*), and adverbials (e.g., *in a much better place*) (Hunston, 2010). This diversity makes it impossible to create a definitive list of lexical items to search for in a corpus. Lastly, another significant challenge in automating the annotation of pragma-discursive features arises from their context-dependent nature. For instance, the statement "we'll arrive by five o'clock" could be understood as a straightforward prediction in certain situations, but in others it should be seen as a prediction that also conveys a promise (Weisser, 2015: 85). Similarly, the word *sorry* does not carry the illocutionary force of an apology when it is used to simply express sympathy with someone else's misfortune or when the apology is mentioned in indirect speech.

[1] Throughout the article, we use the term 'pragma-discursive' to refer to this level of analysis. By using a single term to refer to the annotation of pragmatic and discursive features, we aim to highlight the fact that the challenges involved in their annotation are broadly comparable, as discussed in Section 2.1.

Given the complexities described above, researchers in corpus-based pragmatics have tended to take a form-to-function approach, focusing on well-known lexical markers of illocutionary force, such as politeness formulae and discourse markers (O'Keeffe, 2018; Weisser, 2016). This approach is also prevalent in corpus-assisted discourse studies (CADS), where communicative functions are often investigated by looking at a limited set of reliable lexical indicators. For example, Baker et al. (2019) analyse how evaluation is expressed in patients' comments about the UK's National Health System by looking at the 10 most frequent positive and negative words found in the corpus. While this approach has the benefit of being replicable and scalable, it is likely to miss at least some of the ways in which evaluative meanings can be conveyed. This is especially the case when evaluative expressions span multiple words and when opinions and feelings are invoked via less explicit language (Martin & White, 2005).

While form-to-function approaches remain prevalent in both corpus pragmatics and CADS, a growing body of work is shifting towards a function-to-form approach. This approach involves manual corpus annotation, focusing on the specific pragmatic or discursive phenomenon being studied rather than a predetermined set of lexical units. Several pragmatically annotated corpora have been created to study features such as speech acts (Kirk, 2016; Milà-Garcia, 2018), im/politeness (Taylor, 2016) and advice (Põldvere et al., 2022). Within CADS, discourse features that have been explored through manual corpus annotation include appraisal (e.g., Cavasso & Taboada, 2021), constructiveness and toxicity (Kolhatkar et al., 2020), stance (e.g., Simaki et al., 2020), metaphor (e.g., Fuoli et al., 2022) and rhetorical moves (e.g., Yu, 2022). Manual corpus annotation offers two key benefits in comparison to form-driven analysis. First, it allows researchers to identify and consider all instances of a given pragma-discursive phenomenon, regardless of their lexical complexity. Second, it is contextually sensitive because human coders read the corpus texts as they annotate them and can thus make more accurate and nuanced interpretations. However, manually annotating a corpus demands significant resources, which hampers the method's scalability and practicality. For instance, Fuoli and Hommerberg (2015) reported spending on average an hour per 1000 words, despite using a relatively simple coding scheme. Moreover, manual annotation inherently involves subjectivity and is susceptible to inconsistencies and errors stemming from factors like distraction or cognitive fatigue. To bolster reliability, inter-coder agreement tests may be conducted. However, these tests increase to the overall workload, as extensive training of collaborators is needed. As a result, functionally annotated corpora tend to be relatively small, which inevitably raises issues concerning the generalizability of conclusions drawn from them.

To fully unlock the potential of function-to-form analysis, we need computational tools capable of autonomously annotating pragma-discursive features within corpora with a high degree of precision. Some headways have been made in the task of automatic speech act tagging. Weisser's (2016) dialogue annotation and research tool (DART), for example, is reportedly capable of achieving human-level accuracy (Weisser, 2016: 386). The tool employs an algorithm that identifies lexical patterns typically associated with a wide variety of speech acts based on an in-built thesaurus, and combines syntactic and semantic information to automatically infer the speech act performed in each unit of discourse. In the field of NLP, speech act annotation, commonly referred to as 'Dialogue Act Recognition', is a well-established task with several approaches that have shown good accuracy scores (Zhao & Kawahara, 2019). While these tools undoubtedly represent an important step forward, their functionality is limited to the task of speech act analysis. Developing similar tools for other annotation tasks, such as the analysis of apologies based on the local grammar framework, is a considerable undertaking which requires both specialized linguistic knowledge and advanced programming skills. In contrast, LLMs can be instructed using natural language prompts, making them potentially more accessible to a broader spectrum of researchers with varying levels of computational expertise. Therefore, in this study our primary aim is to explore the potential of using LLMs to assist with pragma-discursive annotation, using the local grammar of apology as a test case. If this approach proves viable, it could pave the way for a new phase of large scale, function-to-form research in both corpus pragmatics and CADS.

### **2.2** LLM-assisted corpus annotation

Large language models are reshaping computational approaches to language. These systems leverage data to grasp intricate language structures, enabling human-like text generation, question answering, summarization, and many other language-related tasks. Tools such as ChatGPT and the Bing chatbot offer a conversational interface to a LLM, allowing users to interact with and direct the system through natural language prompts.

LLMs are causing a paradigm shift in the field of NLP, where they are being used for a wide range of tasks and consistently achieving state-of-the-art performance (Yang et al., 2023). Among these applications, the task of text annotation has garnered considerable attention. This interest arises from the fact that many NLP systems are 'trained' on manually annotated datasets. That is, the systems learn from hand-coded data, identifying patterns and adapting their internal mechanisms to better understand and process similar data in the future. Therefore, finding an efficient and cost-effective way to annotate texts is a critical goal in NLP. LLMs have demonstrated impressive abilities to mimic human behaviour, including inference and contextual understanding, making them suitable for data annotation tasks (Yang et al., 2023). Recent research on LLM-assisted annotation in NLP has shown promising results. For instance, in a study by Frei & Kramer (2023), LLMs performed reasonably quite well in a named entity recognition (NER) task consisting of automatically identifying mentions of drugs, their strength, and diagnoses in German medical texts. Ding et al. (2023) assess LLMs across common NLP tasks like sentiment analysis, relation extraction, and NER. They find that these systems can achieve human-level accuracy but at a fraction of the cost. Similarly, Gilardi et al. (2023) demonstrate that ChatGPT outperforms crowd workers in tasks such as relevance, stance, topics, and frame detection. While these results show potential, the extent to which LLMs can produce accurate annotations for the specific phenomena of interest to researchers in corpus pragmatics and CADS remains unclear. Until now, empirical research assessing their capabilities has primarily focused on annotation tasks utilized in NLP. These tasks tend to be practical in nature and rely on analytical frameworks somewhat different from those more commonly employed in corpus linguistics.

An essential part of using LLMs for corpus annotation and other language tasks is crafting specific verbal instructions that condition the model to generate desired outputs accurately. This process is referred to as prompt engineering or 'prompting' and is a rapidly growing focus in AI research (Liu et al., 2023). Several prompting strategies have been developed and tested across a variety of NLP tasks, including text annotation. One of the most basic prompting strategies is *zero-shot* prompting, in which the model only receives a description of the task and must rely on its overall understanding capabilities to generate the output (Brown et al., 2020)[2]. A common alternative technique is *few-shot prompting*, where the model is exposed to a handful of task-related examples ('shots') along with their expected outputs (Brown et al., 2020)[3]. Several more complex prompting strategies have also been tested. For example,

[2] An example of *zero-shot* prompt could be:
Annotate the act of apology in the following text: "sorry, could you close the door?"
[3] An example of *few-shot* prompt could be:
Question: Annotate the act of apology in the following text: "sorry, could you close the door?"
Answer: <APOLOGY> sorry </APOLOGY>, could you close the door?
Question: Annotate the act of apology in the following text: "oh sorry mum there you go ok"

He et al. (2023) use *few-shot chain-of-thought* prompting, which involves using a series of logically connected prompts, each building on the previous one, to simplify the annotation task by breaking it into smaller steps and guide the model’s responses. Specifically, He et al. (2023) provide ChatGPT with category definitions, followed by a set of labelled examples with concise justifications for each. In a similar vein, Wei et al. (2023) propose a two-stage prompting framework for NER. The first stage uses question-answer dialogue with ChatGPT to identify the types of entities, relations, or events found in a sentence, helping to narrow down the classification task. In the second stage, the model is asked to match the categories found in the sentence with their corresponding words. Recognizing the centrality of prompting in optimizing the performance of LLMs, in this study we draw inspiration from the work reviewed here to develop a tailored strategy for annotating apologies. By doing so, we offer a replicable protocol that could be either applied at a larger scale for the same annotation task or adapted to similar pragmatic and discourse-level annotation tasks.

## 3. Data and methods

To assess the viability and accuracy of AI-assisted pragma-discursive annotation, we employ the GPT-3.5 and GPT-4 to code the functional components of apologies based on a local grammar approach (Su and Wei, 2018). This section outlines the procedure we have developed for this task. The approach consists of three main steps: 1) defining the annotation task, 2) designing a prompt that enables the selected LLMs to generate the desired annotated outputs, and 3) evaluating the performance of the LLMs.

### **3.1** Defining the annotation task

Our experiment focuses on the task of local grammar annotation. Local grammar is an approach to linguistic analysis that seeks to describe the lexico-grammatical patterns associated with a specific meaning or function (Hunston, 2002: 178). As demonstrated in previous research, the local grammar approach is particularly useful for analysing pragmatic functions, or more specifically speech acts such as evaluation (Hunston &

---

Answer: oh <APOLOGY> sorry </APOLOGY> mum there you go ok”
Question: Annotate the act of apology in the following text: “Oh sorry darling I ‘m not running off with you.”
Answer: ___________

Sinclair, 2000; Hunston & Su, 2019), request (Su, 2017), apology (Su & Wei, 2018), disclaiming (Cheng & Ching, 2018), and exemplification (Su & Zhang, 2020).

The analysis of speech acts based on the local grammar framework typically involves five key steps. First, we identify the lexical markers that conventionally realize the speech act of interest drawing on previous work and exploratory corpus analysis. Second, we build a sub-corpus of utterances containing these lexical markers. Next, we analyse a sample from the sub-corpus qualitatively to identify the core functional elements that constitute the target speech act. This analysis is formalized into a codebook which will guide the annotation process. Fourth, we manually annotate all the utterances in the sub-corpus according to the codebook. Finally, we analyse the annotated corpus to uncover the local grammar patterns of the speech act. LLMs can be used to improve the efficiency and reduce the workload needed to carry out the fourth step of the local grammar analysis procedure. It is important to note that the initial step of this procedure relies on a form-to-function approach, whereas the fourth step takes a function-first perspective, as the lexical realizations of the functional components of a speech act are not predetermined. The integration of form-first and function-first approaches makes local grammar analysis an ideal testing ground for evaluating the capabilities of LLMs in corpus coding. This approach allows us to not only assess the model's ability to accurately annotate lexical realizations of open-ended functional categories (i.e., the functional elements of a speech act) but also to gauge its accuracy in coding lexical markers that are known to carry a given illocutionary force in some but not all contexts (e.g., as discussed above, *sorry* may not always convey an apology).

In this study, we focus on analysing the local grammar of the speech act of apology in English, which has been previously investigated in several studies (e.g., Su, 2021; Su & Wei, 2018). We select the apology speech act to test our approach because it has received considerable attention in corpus pragmatics and effectively illustrates the common challenges in annotating functional discourse elements, as discussed above. As an exploratory, 'proof of concept' study, our examination is limited to apology utterances that were expressed using the word *sorry*, which is the most commonly used lexical marker for apologising in English (Lutzky & Kehoe, 2017b). The corpus used for the analysis includes 5,539 instances containing *sorry* extracted from the Spoken BNC2014 (Love et al., 2017), which contains real-life informal conversations between speakers of British English from across the United Kingdom. Each instance is 20 tokens in length.

According to Su and Wei (2018), the seven functional elements that are frequently associated with explicit apologies in English are: APOLOGISER (the individual who apologises), APOLOGISING (the word or expression that realizes the apology, equivalent to the IFID), FORGIVENESS-SEEKING (the action of seeking forgiveness), APOLOGISEE (the recipient of the apology), INTENSIFIER (expressions intensifying the level of regret), SPECIFICATION (specifying the offense or reason for the apology), and HINGE (grammatical devices linking different functional elements). For example, an apology can be construed as the pattern "APOLOGISER + HINGE + APOLOGISING" (e.g., I'm sorry), or with the pattern "FORGIVENESS-SEEKING + APOLOGISEE" (e.g., *Forgive me, John*). Since LLMs' ability to perform functional annotation on apology utterances had not been previously tested, we chose to begin with a simpler, streamlined framework. In consequence, the element of FORGIVENESS-SEEKING was not included in our annotation scheme as it is primarily realized through lexical markers such as *forgive* and *pardon*, which were not used as search terms to build our sample. The element of HINGE, which refers to the copula *BE* and other closed class words such as *for* and *that*, was not considered as it was deemed less central to our experiment and may potentially confuse LLMs. Finally, we decided to rename the element of SPECIFICATION to REASON, which we expected the LLMs to be able to understand more easily. In summary, the annotation task involved two key steps: (1) identify the apology speech act, and (2) annotate apologies by using terms including APOLOGISING, REASON, APOLOGISER, APOLOGISEE, and INTENSIFIER.

### **3.2** Prompt design

As discussed above, a crucial aspect of utilizing and optimizing the performance of LLMs is the creation of well-crafted natural language prompts. The process of prompt engineering generally follows a progressive trial-and-error approach. Various strategies for instructing the model are tested, and the outcomes serve as a guide for refining the prompt. In this study, we employed the Bing chatbot (with the precise mode) for prompt design and testing. We crafted an initial prompt and assessed its effectiveness across three samples, each comprising 100 instances containing *sorry*. In the initial two rounds of testing, we systematically refined the prompt to enhance the annotation performance of the Bing chatbot. In the final round of testing, the fine-tuned prompt yielded annotated results with an accuracy rate of 98%, demonstrating its suitability for our annotation task.

The version of the prompt used in our experiment is shown in Appendix 1. The technique we used is few-shot prompting, in which task instructions are enriched with a set of examples to help the LLM 'understand' the logic underlying the task at hand. Accordingly, our prompt is made up of three components: category definitions, a set of annotated exemplars and the task instructions, formulated as a question. Due to the token limit of the Bing chatbot, the prompt was divided into two parts, which had to be input separately.

We chose the exemplars to be included in the prompt based on the following criteria:

(i) Representativeness. The chosen exemplars contain highly frequent phraseology of the word *sorry*. To identify frequent phraseological patterns, we used AntConc to extract 2-gram, 3-gram, and 4-gram clusters (with a frequency of ≥10) such as *I'm sorry*, *sorry I've, really sorry*, *sorry about that*. These clusters were then searched within the corpus to retrieve instances that could serve as exemplars in the prompt.

(ii) Diversity. The exemplars were selected to encompass various local grammar patterns, capturing the diversity of apology expressions.

(iii) Conciseness. Our preliminary tests revealed that the more concise the prompt, the better the LLM's performance in the annotation task. Therefore, we excluded redundant exemplars.

During the process of designing and testing prompts, we made adjustments to the exemplars to improve Bing chatbot's performance. When analysing test samples, if Bing chatbot repeatedly made specific errors, we introduced relevant exemplars to help it avoid those mistakes. For instance, we included exemplars with the REASON functional tag, which proved to be more challenging for the machine to identify compared to other tags. The final prompt consisted of 10 carefully chosen exemplars. In addition to selecting appropriate exemplars, we also observed that various other factors impacted the LLM's performance, as outlined below:

(i) Formal layout. The use of textual boundaries, such as paragraph segments and the Q&A format, enhanced the LLM's understanding of the prompt.

(ii) Grammatical correctness. Eliminating grammatical errors in the corpus examples enhanced the LLM's performance.

(iii) Terminological precision. Using precise terminology in the prompt improved the LLM's performance compared to using general or vague words. For example, the machine better understood the expression of *the speech act of apology* than the more general expression of *the utterance of apology*.

(iv) Explicitness. The machine generated more accurate results when we specified the types of functional elements to be identified in the final question (*Can you detect the speech act of apology and annotate any functional elements such as* APOLOGISER, REASON, APOLOGISEE, APOLOGISING, *or* INTENSIFIER *in the following utterance?*).

(v) Textual conciseness. Both instructions and exemplars were kept as concise as possible since complex texts could 'confuse' the machine.

(vi) Textual order. The order of textual units influenced the machine's attention priority. In our annotation task, one major difficulty regards the functional element of REASON. In the process of prompt testing, we found that moving the exemplars containing the functional element of REASON to the beginning of the prompt improved the machine's performance in annotating the specific element.

(vii) Label clarity. Category labels that are semantically transparent, explicit, and less technical were better understood by the LLM. For example, replacing the tag SPECIFICATION with the tag REASON enhanced Bing chatbot's annotation performance.

(viii) Inappropriate language. In cases where the texts to be annotated contained sensitive or inappropriate language (e.g., *bitch*), the Bing chatbot was unable to generate texts reproducing those words due to its content moderation filters.

By considering these aspects and making appropriate adjustments, we developed a prompt that effectively guided the Bing chatbot to complete the annotation task for this study.

**3.3** Performance evaluation

The performance of LLMs on the task of local grammar annotation was assessed in two stages. First, we compared two of the most advanced and widely used LLMs, GPT-4, which powers the Bing chatbot, and GPT-3.5, underpinning ChatGPT, to determine the most suitable model for our task. We evaluated their performance on a sample of 50 instances retrieved from our corpus. Once we identified the best performing LLM, we proceeded to compare it with a human annotator. This comparison involved annotating 1000 instances from our corpus. At this stage, one of the authors, serving as an assessor, evaluated the annotated results. The assessor took into consideration both the definitions of the tags and the expanded co-text in each instance to determine whether a textual unit was accurately annotated and whether a tag was adequately assigned. In a few cases, a degree of subjectivity may still unavoidably arise due to the lack of comprehensive contextual information related to a pragmatic unit.

## 4. Results

In this section, we firstly present a comparative performance analysis between the GPT-3.5/ChatGPT and GPT-4/Bing. Next, we show the results of the comparison between the best performing LLM and a human annotator.

**4.1** GPT-3.5 versus GPT-4

The experiments were carried out using the ChatGPT chatbot and Bing chatbot (with theprecise mode), accessible at https://chat.openai.com/ and https://www.bing.com/new, respectively. First, we entered an identical prompt (Appendix A) into both chatbots. Next, we provided both chatbots with a randomly selected sample of 50 instances from our corpus as input. Lastly, we gathered and evaluated the accuracy of the annotated instances derived from the generated output text. An instance was considered as accurately annotated only when the pragmatic elements that it contained were all correctly coded. The results, presented in Table 1, show that GPT-4/Bing clearly outperformed GPT-3.5/ChatGPT at the instance-level.

Table 1. Instance-level accuracy obtained with the two LLMs tested

| | GPT-4/Bing | GPT-3.5/ChatGPT |
|---|---|---|
| No. of tested instances | 50 | 50 |
| No. of correctly annotated instances | 42 | 25 |
| Instance-level accuracy (%) | 84 | 50 |

GPT-3.5/ChatGPT's comparatively poor performance can be largely attributed to the following issues: 1) confusion in tag assignment, such as annotating *sorry* as APOLOGISER instead of APOLOGISING; 2) misidentification of tags like REASON, INTENSIFIER, and others; and 3) inconsistencies in the annotation format, where generated texts deviated from the prompts. Examples of these inaccuracies are shown in Appendix B. Recognizing that GPT-3.5/ChatGPT's subpar performance may be linked to inadequate prompts, we undertook additional efforts to identify more effective prompting strategies. Ultimately, we discovered that GPT-3.5/ChatGPT could more accurately identify the functional elements of apologies when extraneous text irrelevant to the apology speech act was excluded in advance. In other words, to effectively conduct the annotation task with GPT-3.5/ChatGPT, we would need to design another prompt for the model to pre-process and 'sanitize' the corpus instances in order to strip of text fragments that do not pertain specifically to the speech act of apology. However, the addition of the pre-processing step is not ideal because it would complicate the annotation procedure and may introduce potential noise to the data to be annotated (e.g., retaining linguistic parts that do not really indicate the reason for an apology). Consequently, we decided to shift our focus towards evaluating GPT-4/Bing's performance on a larger sample, comparing it to human annotation.

### **4.2** GPT-4 versus a human annotator

A set of 1000 randomly selected instances from the corpus served as the basis for assessing the annotation capabilities of GPT-4/Bing in comparison to those of a human annotator. For the evaluation of the GPT-4/Bing chatbot's performance, we processed each of the 1000 instances individually, as we found that inputting multiple instances at once yielded less accurate results. The annotation task was carried out between April 11th and April 28th, 2023. An example of the annotated output from Bing can be found in Appendix C. To create the annotated dataset for comparison, we recruited a human annotator and tasked them to manually code the same set of examples. We provided the annotator with instructions similar to those the chatbot received. Next, we asked the annotator to use the Note Tab program to conduct three annotation tests on three samples, each containing 100 instances randomly selected from the corpus. These tests were conducted to ensure that the annotator understood the task properly and to improve agreement between the annotator and the task assessor. Once the annotator achieved 100% accuracy on the third set of instances, they proceeded to code the full dataset of 1000 instances. As reported by the annotator, it took approximately 4 hours to complete the task. The annotated texts generated by the Bing chatbot and the human annotator were then evaluated by the assessor.

Following standard practice in NLP, we used *precision*, *recall* and *F1-score* to evaluate and compare the annotation results at the tag level. Precision is a metric that calculates the proportion of accurate positive predictions (true positives) out of all predictions made by the model or human coder (true + false positives). Thus, to take the code REASON as an example, precision answers the question 'out of all the instances that were coded REASON, how many were truly REASON?'. It is computed using the following formula:

$$precision = \frac{TP}{TP+FP} \quad (1)$$

Recall measures the proportion of true positive examples that were correctly identified by the model or human coder. So, to take the code REASON as an example again, recall answers the question 'out of all instances of REASON included in the sample, how many were actually identified?'. It is calculated using the following formula:

$$recall = \frac{TP}{TP+FN} \quad (2)$$

F1-score combines precision and recall into a single value. It represents the harmonic mean of precision and recall, providing an overall assessment of the model's accuracy and coverage for a particular category. The formula for F1-score is:

$$F_1 = \frac{2*precision*recall}{precision+recall} \quad (3)$$

As shown in Table 3, GPT-4/Bing achieved high levels of accuracy at the instance-level, although its performance fell slightly short of that of the human coder. Nevertheless, the chatbot outperformed the human coder by a slight margin when it came to annotating the functional element of REASON. In the sections below, we review the performance of both GPT-4/Bing and the human coder for each of the categories considered.

Table 2. Accuracy measures for 1000 corpus instances annotated by GPT-4/Bing and the human annotator

| | GPT-4/Bing | | | human annotator | | |
|---|---|---|---|---|---|---|
| Instance-level accuracy (%) | 92.7 | | | 95.4 | | |
| Tag-level performance | Precision (%) | Recall (%) | F1 (%) | Precision (%) | Recall (%) | F1 (%) |
| NO APOLOGY* | 100 | 71.43 | 83.33 | 100 | 88.78 | 94.05 |
| APOLOGISING | 100.00 | 99.91 | 99.95 | 100 | 99.91 | 99.95 |
| REASON | 94.74 | 89.26 | 91.91 | 92.86 | 85.95 | 89.27 |
| APOLOGISER | 91.11 | 100 | 95.35 | 100 | 98.78 | 99.39 |
| APOLOGISEE | 97.22 | 83.33 | 89.74 | 100 | 88.10 | 93.67 |
| INTENSIFIER | 100.00 | 93.18 | 96.47 | 100 | 97.73 | 98.85 |

*This category refers to instances that do not contain the illocutionary force of apology. These instances were treated separately and excluded when calculating the precision and recall rates for each functional tag.

#### **4.2.1** *Recognition of NO APOLOGY*

Instances containing the lexical marker *sorry* do not indicate the presence of a direct speech act of apology in two main scenarios: when *sorry* is used to express sympathy with someone else's misfortune, as in Example 1, and when the apology is mentioned in indirect speech, as in Example 2.

(1) over as er as they thought so I 'm very **sorry** to hear that --ANONnameM yeah still you never know something

(2) she sort of said --UNCLEARWORD and he said I 'm **sorry** I 'm going to have to cut you off there

In our test sample, 98 out of the 1000 instances meet these conditions and consequently should not be annotated. However, the Bing chatbot mistakenly labelled 28 instances as apologies, while the human annotator made 11 errors. Among the 28 instances misclassified by GPT-4/Bing, 20 cases involved apologies mentioned in indirect speech. This inaccuracy may be due to the prompt not clearly specifying that apologies in indirect speech should be considered as NO APOLOGY.

### **4.2.2** *Recognition of* APOLOGISING

In local grammar analysis, the functional element of APOLOGISING can be expressed through lexical markers such as *sorry*, *apologise*, *apologies*, etc. In this particular study, we focused only on instances containing the lexical marker *sorry*. In cases where the speech act of apology is present, each occurrence of the lexical marker *sorry* should be annotated with the tag <APOLOGISING>, as shown in Example 3.

> (3) what ca- the Qatarian? drinking my water oh <APOLOGISING> **sorry** </APOLOGISING> <APOLOGISING> **sorry** </APOLOGISING> I just saw it waving at me I 've got (Annotated by the Bing chatbot)

Out of the 902 instances of apology, both the Bing chatbot and the human annotator correctly annotated all but one case as APOLOGISING, as shown in Example (4) and Example (5). This means that GPT-4/Bing performed exceptionally well in recognizing this functional element which is associated with a fixed form (i.e., *sorry*) in our annotation task.

> (4) lady who works at the Kitchen Garden Café oh so <APOLOGISING> sorry </APOLOGISING> <APOLOGISING> sorry </APOLOGISING> you <APOLOGISEE> **sorry** </APOLOGISEE> yeah go on then apologies so she (Annotated by the Bing chatbot, *sorry* incorrectly coded)
>
> (5) and all that kind of stuff yeah and and it <REASON> **sorry** </REASON> its the first population would have a population of like (Human annotation, *sorry* incorrectly coded)

### **4.2.3** *Recognition of* REASON

The functional element of REASON explains why someone is apologising or what they are apologising for. Recognizing this element requires a strong ability to understand the meaning of words in context. Out of the 121 instances of REASON, the Bing chatbot accurately annotated 108 (see Example 6). Only 13 cases went unnoticed (see Example 7), and 6 cases were mistakenly labelled (see Example 8). These findings underscore GPT-4's impressive capacity to independently discern the reason for an apology and accurately annotate instances of this open-ended functional category, all with minimal examples and without input regarding the specific language markers linked to it.

(6) August birthday --UNCLEARWORD you just had your birthday happy birthday <APOLOGISING> sorry </APOLOGISING> <REASON> **I missed it** </REASON> did you have a good time? (Annotated by the Bing chatbot, REASON correctly identified)

(7) <APOLOGISING> sorry </APOLOGISING> I forgot she was honest that would be lying <APOLOGISING> sorry </APOLOGISING> **I forgot your birthday** you don't mean anything ah (Annotated by the Bing chatbot, REASON not identified)

(8) don't moan at me leave me alone <APOLOGISER > I </APOLOGISER> 'm <APOLOGISING> sorry </APOLOGISING> <REASON> **I 'm just trying to make it nice for you** </REASON> (Annotated by the Bing chatbot, REASON incorrectly identified)

In fact, the results suggest that the human annotator did not outperform GPT-4/Bing in accurately annotating the REASON for apologies. Specifically, there were 17 cases that were overlooked (Example 9), while 8 cases were misclassified (Example 10). It is important to note that these errors do not necessarily imply a deficiency in the annotator's ability to discern the underlying cause of an apology. Rather, they are more likely attributed to cognitive fatigue stemming from the task's complexity and extended duration.

(9) and then --UNCLEARWORD beer when we 're there? --UNCLEARWORD <APOLOGISING> sorry </APOLOGISING> **just landed in on your shoe** would you like to (Human annotation, REASON not identified)

(10) hold on I 've got the digital storytelling up oh <APOLOGISING> sorry </APOLOGISING > <REASON> **I 'll tell you about this** </REASON> one erm it says (Human annotation, REASON incorrectly identified)

#### **4.2.4** *Recognition of* APOLOGISER

The functional element of APOLOGISER refers to the person who is apologising. In English, typical lexical forms for APOLOGISER include the first-person pronouns *I* and *we* (Example 8 and Example 9). GPT-4/Bing successfully identified all 164 cases of APOLOGISER, while the human annotator inadvertently overlooked two cases. However, the AI incorrectly classified 16 cases as APOLOGISER, where the lexical form *I* is actually used as a subject of another speech act (Example 11).

> (11) you stop clicking that pen? it's very annoying <APOLOGISING> sorry </APOLOGISING > <APOLOGISER> **I** </APOLOGISER> 'm not saying the same thing there's people (Annotated by the Bing chatbot)

#### **4.2.5** *Recognition of* APOLOGISEE

The functional element of APOLOGISEE indicates the intended recipient of an apology, as illustrated in Example 12. Similar to the functional element of REASON, this aspect is not expressed through fixed lexical forms and recognizing it requires the machine to have a strong grasp of language in context. The recall rate of GPT-4/Bing for APOLOGISEE is relatively lower compared to APOLOGISER and APOLOGISING, which are more strongly associated with conventional linguistic forms. Out of the 42 cases of APOLOGISEE, 7 cases were not recognized (Example 13), and one case was misclassified. In contrast, the human annotator performed slightly better, overlooking only 5 cases and making no incorrect identifications.

> (12) it well we do not have really any mm? <APOLOGISING> sorry </APOLOGISING> <APOLOGISEE> **darling** </APOLOGISEE> what did you say? I didn't say. (Annotated by the Bing chatbot, APOLOGISEE correctly identified)
>
> (13) whoops ah making a right dog 's dinner of this <APOLOGISING> sorry </APOLOGISING> **man** I 'll have a go have a go well (Annotated by the Bing chatbot, APOLOGISEE not identified)

#### **4.2.6** *Recognition of* INTENSIFIER

INTENSIFIER refers to the element that boosts the intensity or degree of an apology. Out of the 44 instances of INTENSIFIER in our sample, the human annotator missed only one case, while GPT-4/Bing failed to spot three cases. However, the machine's errors do not necessarily indicate a lack of understanding of the meaning of this category. In some instances, the Bing chatbot correctly recognized the linguistic form that was missed in others, as shown in Examples (14) and (15), respectively.

> (14) ? Quite silly nice oh yeah well <APOLOGISER> I <APOLOGISER> 'm <INTENSIFIER> **very** </INTENSIFIER> <APOLOGISING> sorry </APOLOGISING> <REASON> I had to bale </REASON> no probs no problem it did (Annotated by the Bing chatbot, *very* identified as INTENSIFIER)
>
> (15) annoying when I was a little boy <APOLOGISER> I </APOLOGISER> 'm **very** <APOLOGISING> sorry </APOLOGISING> I I hear he's like two little boys I (Annotated by the Bing chatbot, *very* not identified as INTENSIFIER)

**4.2** Summary of findings

To explore the potential of leveraging large language models (LLMs) for automating the annotation of pragma-discursive features in corpora, we conducted a comparative study involving GPT-3.5 (implemented in ChatGPT), GPT-4 (powering the Bing chatbot), and a human annotator. The findings of our study revealed that GPT-4/Bing exhibited superior performance in several key aspects when compared to GPT-3.5/ChatGPT in the task of annotating functional elements associated with the speech act of apology. First, GPT-4/Bing consistently generated output in a more stable manner, whereas GPT-3.5/ChatGPT displayed variability in its responses across different conversation turns. Second, GPT-4/Bing consistently adhered to the specified format for presenting annotated texts as per the given prompts. Third, GPT-4/Bing demonstrated a high degree of accuracy in tag usage, whereas GPT-3.5/ChatGPT exhibited occasional inaccuracies, such as substituting the tag <APOLOGISER> for <APOLOGISING>. Lastly, Bing showed an overall stronger grasp of local grammar tags. Based on these findings, we propose that utilizing GPT-4, implemented in the Bing chatbot, would be a more suitable choice for automating the annotation of local grammar elements.

To determine the extent to which the annotation task can be fully automated or if human intervention remains necessary, we conducted a comparative analysis of the performance between GPT-4/Bing and a human annotator. The results show that the Bing chatbot attained an impressive instance-level accuracy rate of 92.7%, only marginally below the human annotator's accuracy of 95.4%.

When looking at the tag-level performance, both GPT-4/Bing and the human annotator showed differing levels of accuracy. Variation in accuracy was linked to how flexible the linguistic forms representing local grammar functions were. Tags related to highly conventional forms were generally annotated more accurately. For example, for the tag of APOLOGISING realized by *sorry*, both the Bing chatbot and the human annotator achieved an F1 score of 99.95%. For the tag of APOLOGISER, mostly expressed by the first-person pronoun *I*, GPT-4/Bing achieved an F1 score of 95.35%, while the human annotator achieved 99.39%. Nonetheless, the strong connection between function and form occasionally prompted GPT-4/Bing to make overly broad generalizations. In some cases, the Bing chatbot persistently categorized the pronoun *I* as APOLOGISER in irrelevant text fragments that appeared next to the target apology utterance, whereas human annotators did not exhibit this particular error, maintaining a flawless precision rate of 100%.

Functional categories instantiated via a broader and more diverse range of linguistic resources presented challenges for both GPT-4/Bing and the human annotator. Two prime examples are the tags of REASON and APOLOGISEE. GPT-4/Bing achieved F1 scores of 91.91% and 89.74% for these tags, while the human annotator achieved scores of 89.27% and 93.67%, respectively. Despite slight differences, it is important to note that Bing performed well in annotating these tags, demonstrating its strong language comprehension and annotation capabilities.

The most noticeable weakness in GPT-4/Bing's performance concerned the recognition of NO APOLOGY, with a recall rate of only 71.43%. In contrast, the human annotator achieved 88.78%. Specifically, the Bing chatbot tended to 1) misunderstand cases where *sorry* was used to express sympathy towards someone and 2) incorrectly recognize cases where the apology was mentioned in indirect speech. While the former mistake might indicate a weakness in contextual understanding, the latter could potentially be avoided by providing relevant exemplars and clearer instructions in the prompt.

## 5. Conclusion

Annotation is a crucial aspect of contemporary corpus linguistics, helping to capture sophisticated patterns of language structure and use. While certain linguistic features can be automatically annotated relatively easily, pragmatic and discursive elements still require manual annotation due to their complexity and lack of direct and unequivocal mapping onto specific lexical forms. However, manual annotation is time-consuming and prone to errors, which has so far hindered the scalability and potential of function-to-form approaches in corpus linguistics. Against this backdrop, this study set out to explore the possibility of automating the process of pragma-discursive corpus annotation by leveraging the advanced language processing capabilities of LLMs. To this aim, we compared the performance of GPT-3.5, the model powering ChatGPT, GPT-4, the model behind the Bing chatbot, and a human coder in the task of annotating the functional components of apologies in English based on the local grammar framework (Su & Wei, 2018).

Our results show that local grammar analysis is amenable to LLM-assisted annotation and in our case study GPT-4 performed better than GPT-3.5 in the assigned annotation task. This points to the viability of using LLMs to automate local grammar annotation of speech acts and to further develop speech act annotated corpora, which will be a substantial contribution to corpus pragmatics. More importantly, our exploration demonstrates that the overall accuracy of GPT-4 closely approached to that of a human annotator, which suggests that employing LLMs to assist in the task of annotating apologies, and by extension other speech acts, is a viable option. However, for enhanced reliability, human oversight remains necessary. Throughout our study, we noted variability in Bing chatbot's performance depending on the tag type. Generally, tags linked to formulaic linguistic expressions achieved higher accuracy scores, while those that rely on flexible linguistic forms yielded lower scores. Occasionally, the Bing chatbot also exhibited excessive rigidity, erroneously annotating specific forms even when they served a different pragmatic function. Identifying what tags are more likely to require human validation calls for further experimentation with a broader range of annotation tasks. Despite these limitations, however, the overall accuracy scores remain robust, which means that LLMs could be used as a 'first-pass' technique to automatically generate tentative annotations to be validated by a human coder. This would still significantly reduce the overall workload involved in pragma-discursive corpus annotation, making the process significantly more efficient and manageable.

The current study exclusively centred on annotating apologies, and the examples considered were limited to apology expressions featuring the word *sorry*. As explained in Section 3.1, the apology examples we employed for testing our approach included open-ended components such as REASON, and not all occurrences of *sorry* constituted an actual apology, making this an ideal initial testing scenario for LLM-assisted coding. However, it is important to acknowledge that other lexical markers of apology exist and that more research is necessary to ascertain if the strong performance observed here can be replicated with apology utterances containing a wider variety of lexical markers, as well as other speech acts and discursive features. Additionally, it remains to be determined whether these models can maintain their performance when applied to complete texts instead of individual sentences as well as to data other than spoken conversation. Despite these limitations and open questions, our findings are encouraging, especially given the incredibly rapid pace at which the capabilities of LLMs have been developing. We believe this technology could radically transform how we do corpus-based pragmatics and discourse analysis by enabling quantitative function-to-form research on a much larger scale than ever before. We therefore call for further research to explore the potential applications of this approach in the analysis of other pragmatic and discursive phenomena.

A significant focal point of our study revolved around devising effective prompting strategies for the apology annotation task. The prompt we crafted for this specific task holds potential for further enhancement to bolster accuracy, particularly in light of our findings regarding the identification of non-apology instances. Moreover, it can serve as a valuable template for guiding other LLM-assisted corpus annotation endeavours. Our research has demonstrated that a few-shot prompting approach, featuring straightforward category definitions and model responses, can yield robust results. Additionally, we have provided several suggestions about the factors that may influence LLM performance and how to construct linguistic examples that effectively steer the model toward correct answers. Overall, the possibility of using natural language prompts to enable LLMs for specific linguistic annotation tasks makes the LLM-assisted annotation approach highly accessible for linguists without expertise in programming. We hope our study will encourage others to apply LLM-assisted annotation to different pragma-discursive features.

## Appendices

**Appendix A.** Prompt to be input into the chatbots of Bing and ChatGPT

**(First set)**

Please learn the following contents.

The speech act of apology may contain the following functional elements:

APOLOGISING: the element that indicates the act of apologising

REASON: the offense or the reason for the apology

APOLOGISER：the person who apologies

APOLOGISEE: the person to whom the apology is made

INTENSIFIER: the element that upgrades the degree of apology

Here are some examples:

Question: Can you annotate the speech act of apology in the utterance "Ah, I ‘m really sorry for all that."?

Answer: The annotated version is: Ah, <APOLOGISER> I </APOLOGISER> ‘m <INTENSIFIER> really </INTENSIFIER> <APOLOGISING> sorry </APOLOGISING> <REASON> for all that </REASON>.

Question: Can you annotate the speech act of apology in the utterance "Sorry about that, but I ‘ve got to go to work."?

Answer: The annotated version is: <APOLOGISING> Sorry </APOLOGISING> <REASON> about that </REASON>, but I ‘ve got to go to work.

Question: Can you annotate the speech act of apology in the utterance "Hello Mr [gap:name], I ‘m sorry to bother you, my name is Kathy and I represent"?

Answer: The annotated version is: Hello <APOLOGISEE> Mr [gap:name] </APOLOGISEE>, <APOLOGISER> I </APOLOGISER> ‘m <APOLOGISING> sorry </APOLOGISING> <REASON> to bother you </REASON>, my name is Kathy and I represent

Question: Can you annotate the speech act of apology in the utterance "Sorry sorry Mr [gap:name], I moved too quickly for you."?

Answer: The annotated version is: <APOLOGISING> Sorry </APOLOGISING> <APOLOGISING> sorry </APOLOGISING> <APOLOGISEE> Mr [gap:name] </APOLOGISEE>, <REASON> I moved too quickly for you </REASON>

**(Second set)**

Here are some other examples:

Question: Can you annotate the speech act of apology in the utterance "I’m sorry to hear that"?

Answer: No speech act of apology is present in the utterance "I’m sorry to hear that".

Question: Can you annotate the speech act of apology in the utterance "I felt sorry for your loss"?

Answer: No speech act of apology is present in the utterance "I felt sorry for your loss".

Question: Can you annotate the speech act of apology in the utterance "I ‘m sorry that I ‘ve lost it"?

Answer: The annotated version is: <APOLOGISER> I </APOLOGISER> ‘m <APOLOGISING> sorry </APOLOGISING> <REASON> that I ‘ve lost it </REASON>.

Question: Can you annotate the speech act of apology in the utterance "Er, I think there is a tendending now, for them to say, oh, I ‘m terribly sorry, we can only do this against payment."?

Answer: The annotated version is: Er, I think there is a tendending now, for them to say, oh, <APOLOGISER> I </APOLOGISER> ‘m <INTENSIFIER> terribly </INTENSIFIER> <APOLOGISING> sorry </APOLOGISING>, <REASON> we can only do this against payment </REASON>.

Question: Can you annotate the speech act of apology in the utterance "Oh sorry darling I ‘m not running off with you."?

Answer: The annotated version is: Oh <APOLOGISING> sorry </APOLOGISING> <APOLOGISEE> darling </APOLOGISEE> I ‘m not running off with you.

Question: Can you annotate the speech act of apology in the utterance "oh sorry mum there you go okay"?

Answer: The annotated version is: oh <APOLOGISING> sorry </APOLOGISING> <APOLOGISEE> mum </APOLOGISEE> there you go okay

Question: Can you detect the speech act of apology and annotate any functional elements such as APOLOGISING, REASON, APOLOGISER, APOLOGISEE, or INTENSIFIER in the following utterance? Please exclude any irrelevant texts.

“I’m so excited oh look at these thank you yeah sorry they ‘re a bit wet yeah I like camping that”

**Appendix B**. ChatGPT’s completion (same prompts as those input in Bing’s chatbot)

1. Annotated version: uh we 'll do some more recordings with them um <APOLOGISER> sorry </APOLOGISER>? <REASON> could record them again </REASON> yeah um eastern Christianity puts
2. Annotated version: is it --ANONnameN or --ANONnameN ? yeah --ANONnameN --ANONnameN <APOLOGISER> oh sorry </APOLOGISER> yeah do you remember when I paused it before as
3. Annotated version: yeah exactly --UNCLEARWORD recording --UNCLEARWORD --UNCLEARWORD recording the same time <APOLOGISER> sorry </APOLOGISER> this is this is a flawed system you can watch
4. Annotated version: want another cakey ? another one ? no ? <APOLOGISER> oh sorry </APOLOGISER> okay it 's okay do you want any of that
5. Annotated version: sure you do n't feel like it ? not really <APOLOGISER> sorry </APOLOGISER> fair enough it 's a great --UNCLEARWORD to full body
6. Annotated version: yep erm so --ANONnameM got one one day of work <APOLOGISER> sorry </APOLOGISER> one week of work oh right which he 's using
7. Annotated version: Husband 's Secret okay this is n't a great --UNCLEARWORD <APOLOGISER> sorry </APOLOGISER> I read one of them when I was ill I
8. Annotated version: be changed very much where I live what 's that <APOLOGISER> sorry </APOLOGISER>? I do n't think much has changed in my
9. Annotated version: should be inside --UNCLEARWORD oh my god actually --UNCLEARWORD anything <APOLOGISER> sorry </APOLOGISER>? do you actually feel like anything from there --UNCLEARWORD
10. Annotated version: what does n't taste nice ? your hand ? --UNCLEARWORD <APOLOGISER> sorry </APOLOGISER>? grease on my hand can I go and quickly

**Appendix C.** An example of annotated texts generated by the Bing chatbot

Please learn the following contents. The speech act of apology may contain the following functional elements: APOLOGISING: the element that indicates the act of apologising REASON: the offense or the reason for the apology APOLOGISER: the person who apologies APOLOGISEE: the person to whom the apology is made INTENSIFIER: the element that upgrades the degree of apology Here are some examples: Question: Can you annotate the speech act of apology in the utterance "Ah, I 'm really sorry for all that."? Answer: The annotated version is: Ah, <APOLOGISER> I </APOLOGISER> 'm <INTENSIFIER> really </INTENSIFIER> <APOLOGISING> sorry </APOLOGISING> <REASON> for all that </REASON> . Question: Can you annotate the speech act of apology in the utterance "Sorry about that, but I 've got to go to work."? Answer: The annotated version is: <APOLOGISING> Sorry </APOLOGISING> <REASON> about that </REASON> , but I 've got to go to work. Question: Can you annotate the speech act of apology in the utterance "Hello Mr [gap:name] , I 'm sorry to bother you , my name is Kathy and I represent"? Answer: The annotated version is: Hello <APOLOGISEE> Mr [gap:name] </APOLOGISEE> , <APOLOGISER> I </APOLOGISER> 'm <APOLOGISING> sorry </APOLOGISING> <REASON> to bother you </REASON> , my name is Kathy and I represent Question: Can you annotate the speech act of apology in the utterance "Sorry sorry Mr [gap:name] , I moved too quickly for you."? Answer: The annotated version is: <APOLOGISING> Sorry </APOLOGISING> <APOLOGISING> sorry </APOLOGISING> <APOLOGISEE> Mr [gap:name] </APOLOGISEE> , <REASON> I moved too quickly for you </REASON>

Thank you for sharing this information with me. I have learned the contents and can now annotate the speech act of apology in utterances. Is there anything else you would like me to learn or do?

1 of 20

Here are some other examples: Question: Can you annotate the speech act of apology in the utterance "I'm sorry to hear that"? Answer: No speech act of apology is present in the utterance "I'm sorry to hear that". Question: Can you annotate the speech act of apology in the utterance "I felt sorry for your loss"? Answer: No speech act of apology is present in the utterance "I felt sorry for your loss". Question: Can you annotate the speech act of apology in the utterance "I 'm sorry that I 've lost it"? Answer: The annotated version is: <APOLOGISER> I </APOLOGISER> 'm <APOLOGISING> sorry </APOLOGISING> <REASON> that I 've lost it </REASON>. Question: Can you annotate the speech act of apology in the utterance "Er, I think there is a tendending now, for them to say, oh, I 'm terribly sorry, we can only do this against payment."? Answer: The annotated version is: Er, I think there is a tendending now, for them to say, oh, <APOLOGISER> I </APOLOGISER> 'm <INTENSIFIER> terribly </INTENSIFIER> <APOLOGISING> sorry </APOLOGISING> , <REASON> we can only do this against payment </REASON>. Question: Can you annotate the speech act of apology in the utterance "Oh sorry darling I 'm not running off with you."? Answer: The annotated version is: Oh <APOLOGISING> sorry </APOLOGISING> <APOLOGISEE> darling </APOLOGISEE> I 'm not running off with you . Question: Can you annotate the speech act of apology in the utterance "oh sorry mum there you go okay"? Answer: The annotated version is:oh <APOLOGISING> sorry </APOLOGISING> <APOLOGISEE> mum </APOLOGISEE> there you go okay

Thank you for sharing these examples with me. I have learned that not all utterances containing the word "sorry" contain a speech act of apology. Is there anything else you would like me to learn or do?

2 of 20

Question: Can you detect the speech act of apology and annotate any functional elements such as APOLOGISING, REASON, APOLOGISER, APOLOGISEE, or INTENSIFIER in the following utterance? Please exclude any irrelevant texts. "I'm so excited oh look at these thank you yeah sorry they 're a bit wet yeah I like camping that"

Answer: The annotated version is: "I'm so excited oh look at these thank you yeah <APOLOGISING>sorry</APOLOGISING> <REASON>they 're a bit wet</REASON> yeah I like camping that"

3 of 20